\documentclass[letterpaper, 10 pt, conference]{ieeeconf}
\IEEEoverridecommandlockouts
\usepackage{amsmath}
\usepackage{amssymb}
\usepackage{graphicx}
\usepackage{algorithm}
\usepackage{algorithmic}
\usepackage{multirow}

\title{\LARGE \bf
ReACT: Reflection-Consistent Modulation and Cross-Load Consistency for Humanoid Velocity Tracking Under Asymmetric Loads
}

\author{%
Xianghui Wang$^{1,2}$, 
Aoran Fu$^{3}$, 
Liyou Luo$^{4}$, 
Shanze Wang$^{2,5}$, 
Xiaoyu Shen$^{2}$, 
Dan Zhang$^{1}$, 
Wei Zhang$^{2,\dagger}$%
\thanks{This work has been submitted to the IEEE for possible publication. Copyright may be transferred without notice, after which this version may no longer be accessible.}%
\thanks{*This work was supported by the National Natural Science Foundation of China (Grant No. 62503251).}%
\thanks{$^{1}$The authors are with the Department of Mechanical Engineering, The Hong Kong Polytechnic University, Hung Hom, Kowloon, Hong Kong.
{\tt\small carlos.wang@connect.polyu.hk, dan.zhang@polyu.edu.hk}}%
\thanks{$^{2}$The authors are with the College of Information Science and Technology, Eastern Institute of Technology, Ningbo, China.
{\tt\small \{carlos.wang,shanze.wang\}@connect.polyu.hk, xyshen@eitech.edu.cn, zhw@eitech.edu.cn}}%
\thanks{$^{3}$The author is with the Department of Mechanical Engineering, National University of Singapore, Singapore.
{\tt\small e1499216@u.nus.edu}}%
\thanks{$^{4}$The author is with the Academy of Artificial Intelligence and Advanced Technology, Xi'an Jiaotong-Liverpool University, Suzhou, China.
{\tt\small Lly3215879948@gmail.com}}%
\thanks{$^{5}$The author is with the Department of Aeronautical and Aviation Engineering, The Hong Kong Polytechnic University, Hung Hom, Kowloon, Hong Kong.
{\tt\small shanze.wang@connect.polyu.hk}}%
\thanks{$ \dagger$ Corresponding author: Wei Zhang.}%
}

\begin{document}
\maketitle
\thispagestyle{empty}
\pagestyle{empty}

\begin{abstract}
In recent years, humanoid loco-manipulation has progressed to whole-body grasping, lifting, and transport. A basic requirement underlying these tasks is carrying the object to its destination, which depends on the locomotion controller accurately executing velocity commands from the navigation policy. However, carrying an object in one hand introduces a persistent asymmetric upper-body load that can degrade velocity tracking. Persistent tracking biases can accumulate into position and heading deviations, increasing the risk of collisions and navigation failure. To address this problem, we present ReACT, a framework for accurate velocity tracking under asymmetric upper-body loads. Reflection-consistent disturbance modulation uses the estimated wrench to modulate actor features for load compensation while preserving action correspondence under reflected observations. Cross-load tracking consistency training compares loaded and unloaded conditions under the same velocity commands to promote consistent velocity tracking across loading conditions. Simulation experiments on a Unitree G1 humanoid robot show improved velocity tracking and reduced directional bias under varying asymmetric loads. Real-world experiments demonstrate collision-free goal reaching in all 40 tested navigation trials under unilateral hand loads with a fixed mapless navigation policy.
\end{abstract}

\section{Introduction}
\label{sec:introduction}

In recent years, humanoid loco-manipulation has enabled increasingly complex object interactions~\cite{pmlr-v270-zhang25a,jiang2025wholebodyvla}, including whole-body grasping, lifting, and transport~\cite{zhang2025falcon,byrd2026adaptmanip}. A basic requirement underlying these tasks is carrying the object to its destination while navigating among obstacles~\cite{zhang2025drl, compton2026terrain}, which requires the locomotion controller to accurately track velocity commands throughout the transport phase. Unlike wheeled mobile manipulators on statically stable bases~\cite{fu2024mobile}, a walking humanoid robot must maintain balance while carrying the payload, as support contacts switch during each stride. In practice, objects are commonly carried in one hand, introducing a persistent asymmetric upper-body load that can degrade velocity tracking even when walking remains stable~\cite{lyu2026adaptanalyticaldisturbanceawarepolicy}. Such degraded tracking can manifest as persistent lateral or yaw-rate biases, which accumulate into position and heading deviations and increase the risk of collision or goal-reaching failure during navigation.

Existing methods improve locomotion under external loads from several directions. Dynamics randomization~\cite{peng2018sim} and adversarial disturbance training expose policies to variations in dynamics and external forces to improve robustness~\cite{radosavovic2024real, long2024learning}. FALCON jointly trains upper-body and lower-body policies with a force curriculum, learning force-adaptive loco-manipulation from proprioceptive history~\cite{zhang2025falcon}. ADAPT incorporates analytical disturbance feedback into the locomotion policy, improving velocity tracking and reducing directional drift under asymmetric hand payloads~\cite{lyu2026adaptanalyticaldisturbanceawarepolicy}. Its reported trajectories, however, still show residual lateral drift under larger unilateral loads~\cite{lyu2026adaptanalyticaldisturbanceawarepolicy}. For the same velocity command, different load conditions require different compensatory joint responses. The challenge is to adapt these responses to the estimated disturbance while maintaining accurate velocity tracking with minimal variation across loading conditions.

To address this challenge, we present ReACT, a framework for accurate velocity tracking under persistent asymmetric upper-body loads on humanoid robots. Reflection-consistent disturbance modulation uses the estimated wrench to scale and couple reflection-even and reflection-odd actor features, enabling load-dependent compensation while preserving action correspondence under reflected observations. Cross-load tracking consistency training compares motion responses across loaded and unloaded conditions under the same velocity commands, promoting consistent velocity tracking across loading conditions. Quantitative simulation on a Unitree G1 humanoid robot shows that ReACT achieves accurate velocity tracking and reduced directional bias under varying asymmetric loads. Ablation studies support the effectiveness of both components. Real-robot mapless navigation under asymmetric loads demonstrates collision-free goal reaching in the tested scenarios with a fixed navigation policy.

In summary, our contributions are as follows:
\begin{enumerate}
    \item We propose reflection-consistent disturbance modulation, which uses the estimated external wrench to modulate the policy response for load compensation while preserving action correspondence under reflected observations.

    \item We develop cross-load tracking consistency training, which compares motion responses across loaded and unloaded conditions under the same velocity commands to reduce velocity tracking variation across loading conditions.

    \item We validate ReACT through quantitative simulation, ablation studies, and real-robot mapless navigation on a Unitree G1 humanoid robot, demonstrating accurate velocity tracking in simulation and collision-free goal reaching in the tested real-robot navigation scenarios.
\end{enumerate}
\section{Related Work}
\label{sec:related_work}

\subsection{Humanoid Robot Locomotion under External Loads}
\label{sec:rw_disturbance}

Dynamics randomization~\cite{peng2018sim} and adversarial perturbation training improve locomotion robustness by exposing policies to variations in robot dynamics and external forces~\cite{radosavovic2024real, long2024learning}. Broad randomization, however, can compromise tracking accuracy~\cite{lyu2026adaptanalyticaldisturbanceawarepolicy}. History-based adaptation further conditions locomotion policies on latent environmental information inferred from recent state--action sequences, as demonstrated by RMA and its bipedal extension A-RMA~\cite{kumar2021rma, kumar2022adapting}.

For force-interactive humanoid tasks, FALCON jointly trains separate upper-body and lower-body policies with a force curriculum, learning implicit load compensation from proprioceptive histories while tracking upper-body motion and root velocity~\cite{zhang2025falcon}. Explicit disturbance estimates provide a physically interpretable source of feedback. MOB-Net combines a momentum observer with learned model-uncertainty compensation to estimate whole-body external joint torques without dedicated force/torque sensors~\cite{lim2025mob}.

Concurrent ADAPT~\cite{lyu2026adaptanalyticaldisturbanceawarepolicy} and ReACT both use explicit disturbance feedback to improve velocity tracking under external loads. ADAPT supplies analytical whole-body disturbance estimates as policy inputs. ReACT instead uses a learned wrench estimate for reflection-consistent feature modulation and explicitly compares motion responses across matched load conditions to reduce cross-load tracking variation.

\subsection{Symmetry in Humanoid Robot Locomotion}
\label{sec:rw_symmetry}

Morphological symmetries provide structural priors for robot dynamics and learned controllers~\cite{apraez2025morphological}. Data augmentation and symmetry regularization encourage consistent behavior across mirrored locomotion experiences, without guaranteeing exact equivariance~\cite{abdolhosseini2019learning,mittal2024symmetry}. Architectural methods enforce this correspondence by construction. SE-Policy embeds strict reflection equivariance in the actor and invariance in the critic, improving velocity tracking and motion coordination on a real humanoid robot~\cite{nie2026coordinated}. SWAP extends symmetry constraints to the world model and actor--critic networks, improving generalization to mirrored terrains in quadruped parkour~\cite{lan2026swap}.

These symmetry constraints establish correspondence between mirrored observations and actions. Such correspondence alone does not ensure consistent tracking across different loads under the same velocity command, so an equivariant policy can still retain load-dependent tracking bias. ReACT uses the reflection properties of the estimated wrench to scale and couple reflection-even and reflection-odd actor features while preserving this correspondence, and explicitly compares loaded and unloaded motion responses during training.
\section{Preliminaries}
\label{sec:preliminaries}

\subsection{Problem Definition}
\label{sec:problem_definition}

We consider velocity tracking for a humanoid robot subject to asymmetric upper-body loading and external disturbances. At each control step, the policy~$\pi$ selects an action~$a_t$ based on the observation~$o_t$. We seek to maximize the expected discounted return
\begin{equation}
    \pi^{*}
    =
    \arg\max_{\pi}
    \mathbb{E}_{\pi}
    \left[
        \sum_{t=0}^{\infty}\gamma^{t}r_t
    \right],
    \label{eq:rl_objective}
\end{equation}
where~$r_t$ is the reward at step~$t$ and $\gamma\in[0,1)$ is the discount factor.

The locomotion policy receives a velocity command $c_t=(v_x^{\mathrm{cmd}},\,v_y^{\mathrm{cmd}},\,\omega_z^{\mathrm{cmd}})$, where $v_x^{\mathrm{cmd}}$ and $v_y^{\mathrm{cmd}}$ are the desired body-frame forward and lateral velocities, and $\omega_z^{\mathrm{cmd}}$ is the desired yaw rate. The absolute tracking errors in the same frame are
\begin{equation}
\left\{
\begin{aligned}
    e_x      &= \left|v_x-v_x^{\mathrm{cmd}}\right|,\\
    e_y      &= \left|v_y-v_y^{\mathrm{cmd}}\right|,\\
    e_\omega &= \left|\omega_z-\omega_z^{\mathrm{cmd}}\right|.
\end{aligned}
\right.
\label{eq:tracking_error}
\end{equation}
Our objective is to reduce these tracking errors while maintaining stable locomotion.

\subsection{Observation and Action Spaces}
\label{sec:observation_action}

\noindent\textbf{Observation Space.}
The actor receives a 106-dimensional observation $o_t\in\mathbb{R}^{106}$ given by
\begin{equation}
    o_t =
    [
    \omega_t^{b},\,
    g_t^{b},\,
    c_t,\,
    q_t-q^{0},\,
    \dot q_t,\,
    \bar{a}_{t-1},\,
    \Phi_t,\,
    \hat{\mathbf{d}}_t
    ],
    \label{eq:actor_observation}
\end{equation}
where the individual components and their sagittal reflection transformations are summarized in Table~\ref{tab:observation_action}. Quantities with superscript~$b$ are expressed in the robot body frame. The estimated wrench~$\hat{\mathbf{d}}_t$ is expressed about the torso origin in the torso frame, relative to its calibration reference, and is provided by a learned estimator described in Sec.~\ref{sec:impl}. During training, the critic receives an augmented 109-dimensional observation that additionally includes the simulated base linear velocity~$v_t^{b}$.

\noindent\textbf{Action Space.}
The actor outputs a 15-dimensional action $a_t\in\mathbb{R}^{15}$ for six joints in each leg and three waist joints. Under position control, the policy specifies residual joint-position targets relative to the default configuration~$q^{0}$ as
\begin{equation}
    q^{\mathrm{tar}}_{t,\mathcal{J}}
    =
    q^{0}_{\mathcal{J}}+\alpha_a a_t,
    \label{eq:action_target}
\end{equation}
where $\mathcal{J}$ denotes the 15 actor-controlled joints and $\alpha_a$ is the action scale. The 14 arm joints are driven by externally prescribed trajectories and are excluded from the actor action space. Their joint states and previous commands remain part of the actor observation.

The joint-reflection operator~$F_q$ in Table~\ref{tab:observation_action} exchanges corresponding left and right joints with sign changes determined by their rotation axes, and applies identically to the previous full-body command~$\bar{a}_{t-1}$. $F_a$~is the restriction of~$F_q$ to the actor-controlled joints~$\mathcal{J}$.
\section{Method}
\label{sec:method}

\begin{figure*}[t]
    \centering
    \includegraphics[width=0.85\textwidth]{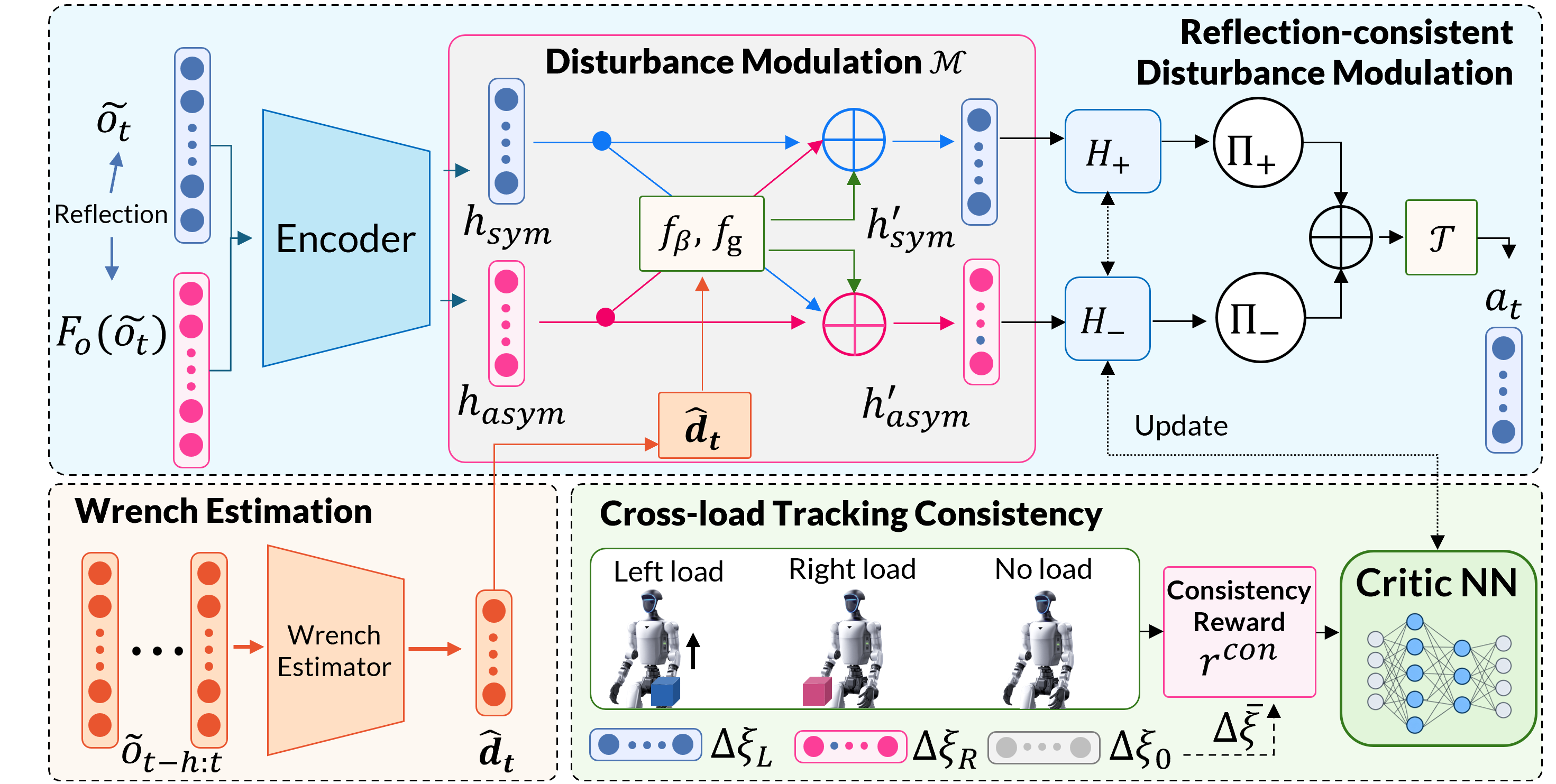}
    \caption{Overview of ReACT.}
    \label{fig:framework}
\end{figure*}

To address velocity tracking degradation under asymmetric upper-body loading, we propose ReACT. As illustrated in Fig.~\ref{fig:framework}, ReACT consists of two components. Reflection-consistent disturbance modulation~(Sec.~\ref{sec:c1}) uses the estimated wrench to modulate actor features for load compensation, maintaining consistent actions under reflected observations. Cross-load tracking consistency~(Sec.~\ref{sec:c2}) promotes consistent command execution across loading conditions through matched-group
response comparison during training.

\subsection{Reflection-Consistent Disturbance Modulation}
\label{sec:c1}

Considering sagittal reflection of humanoid robots, reflecting a hand-held load to the opposite hand preserves the vertical force and reverses the roll moment. Accordingly, we separate the estimated wrench into invariant components $\hat{\mathbf d}_{\mathrm{sym}}=(\hat F_x,\hat F_z,\hat M_y)$ and sign-reversing components $\hat{\mathbf d}_{\mathrm{asym}}=(\hat F_y,\hat M_x,\hat M_z)$. We extend the observation reflection~$F_o$ of SE-Policy~\cite{nie2026coordinated} to reverse $\hat{\mathbf d}_{\mathrm{asym}}$. The architecture described below enforces $\pi_\theta(F_o(o_t))=F_a\pi_\theta(o_t)$ for the deterministic actor, where~$F_a$ is the action reflection.

Specifically, the encoder~$\mathcal E$ receives~$\tilde o_t$, the observation~$o_t$ with wrench components replaced by zero. The wrench is expressed relative to its calibration reference, with reflection-compatible normalization omitted from the notation. The estimated wrench~$\hat{\mathbf d}_t$ enters the actor through the modulation function~$\mathcal M$. The encoder output is decomposed as
\begin{equation}
\begin{aligned}
h_{\mathrm{sym}}
&=\tfrac12\bigl[\mathcal E(\tilde o_t)
+\mathcal E(F_o(\tilde o_t))\bigr],\\
h_{\mathrm{asym}}
&=\tfrac12\bigl[\mathcal E(\tilde o_t)
-\mathcal E(F_o(\tilde o_t))\bigr].
\end{aligned}
\label{eq:decomposition}
\end{equation}
Under reflection, $h_{\mathrm{sym}}$ is invariant and $h_{\mathrm{asym}}$ changes sign.

For~$h=(h_{\mathrm{sym}},h_{\mathrm{asym}})$, each modulation block adjusts feature magnitudes and generates inter-branch updates from the estimated wrench. We collect reflection-invariant quantities into~$\eta$ and compute scale coefficients~$(\beta_{\mathrm{sym}},\beta_{\mathrm{asym}})$ and a sign-reversing coupling quantity~$g$ in~(\ref{eq:modulation}).
\begin{equation}
\begin{aligned}
\eta
&=[h_{\mathrm{sym}},\hat{\mathbf d}_{\mathrm{sym}},
\iota(\hat{\mathbf d}_{\mathrm{asym}})],\\
\eta_0&=[h_{\mathrm{sym}},\mathbf0,\mathbf0],\\
(\beta_{\mathrm{sym}},\beta_{\mathrm{asym}})
&=f_\beta(\eta)-f_\beta(\eta_0),\\
g&=f_g(\eta)\,\hat{\mathbf d}_{\mathrm{asym}}.
\end{aligned}
\label{eq:modulation}
\end{equation}
The function~$\iota$ maps~$\hat{\mathbf d}_{\mathrm{asym}}$ to invariant features comprising absolute values, squares, and pairwise products, satisfying~$\iota(-z)=\iota(z)$. The learned functions~$f_\beta$ and~$f_g$ produce scale parameters and a gain matrix, respectively, from reflection-invariant inputs. The inter-branch terms allow the actor to distinguish~$\hat{\mathbf d}_{\mathrm{asym}}$ from~$-\hat{\mathbf d}_{\mathrm{asym}}$ at a fixed encoder input, which invariant scaling alone cannot achieve. At the wrench reference, $\eta$ reduces to~$\eta_0$.

The scale coefficients and coupling quantity update the features in~(\ref{eq:feature_update}).
\begin{equation}
\begin{aligned}
h'_{\mathrm{sym}}
&=[1+\alpha\tanh(\beta_{\mathrm{sym}})]\odot h_{\mathrm{sym}}\\
&\quad+\mathcal B_{\mathrm{sym}}(g,h_{\mathrm{asym}}),\\
h'_{\mathrm{asym}}
&=[1+\alpha\tanh(\beta_{\mathrm{asym}})]\odot h_{\mathrm{asym}}\\
&\quad+\mathcal B_{\mathrm{asym}}(g,h_{\mathrm{sym}}).
\end{aligned}
\label{eq:feature_update}
\end{equation}
The scale factor~$[1+\alpha\tanh(\beta)]$ equals one when~$\beta=0$, with~$\alpha=0.25$ and~$\odot$ denoting element-wise multiplication. The bilinear maps~$\mathcal B_{\mathrm{sym}}$ and~$\mathcal B_{\mathrm{asym}}$ combine~$g$ with features from the opposite branch through bias-free linear projections and element-wise products. These produce invariant updates for~$h_{\mathrm{sym}}$ and sign-reversing updates for~$h_{\mathrm{asym}}$. At the zero-wrench reference, the scale coefficients and~$g$ vanish, yielding~$h'=h$.

We apply two independently parameterized modulation blocks sequentially, each constructing~$\eta$ and~$\eta_0$ from its input features. The final~$h'$ determines the actor's mean branch. Equation~(\ref{eq:action_mean}) defines its deterministic output for evaluation and deployment, while FastSAC~\cite{seo2025learning} samples stochastic actions during training.
\begin{equation}
\begin{aligned}
u_\theta
&=\Pi_+H_+(h'_{\mathrm{sym}})
+\Pi_-H_-(h'_{\mathrm{asym}}),\\
\pi_\theta(o_t)&=\mathcal T(u_\theta).
\end{aligned}
\label{eq:action_mean}
\end{equation}
Here, $\Pi_\pm=\tfrac12(I\pm F_a)$ with $F_a^2=I$. $H_+$ is affine and $H_-$ is bias-free linear. $\mathcal T$ applies symmetric clamp, element-wise tanh, and per-joint affine scaling, and commutes with~$F_a$. The even head output is projected onto the~$F_a$-even subspace, the odd head is bias-free, and the post-tanh affine offset is zero. Network hyperparameters are specified in Sec.~\ref{sec:impl}.

\subsection{Cross-Load Tracking Consistency}
\label{sec:c2}

Reflection-equivariant actions can still produce load-dependent tracking biases. We therefore regularize the motion responses of matched rollouts under the same velocity command. Specifically, we train the same policy in matched groups of unloaded, left-loaded, and right-loaded branches~$b\in\mathcal{I}=\{\mathrm{U},\mathrm{L},\mathrm{R}\}$. The branches start from matched robot states at group reset and execute shared velocity commands throughout each comparison window.

To assess accumulated differences in command execution, we compare planar motion responses over windows of duration~$T$. For each branch, $\Delta\xi_b\in\mathbb{R}^3$ comprises the net horizontal displacement of the robot root, expressed in that branch's starting yaw frame, and the heading change wrapped to~$[-\pi,\pi)$.

For groups completing the window without failure, the consistency reward in~(\ref{eq:consistency_reward}) is added at window completion, penalizing deviations from the group mean~$\bar{\Delta\xi}$.
\begin{equation}
\begin{aligned}
\zeta_b
&=\bar{\Delta\xi}-\Delta\xi_b,\\
r_b^{\mathrm{con}}
&=-\lambda_{\mathrm{con}}
\min\!\left\{c_{\max},\,
\bar\rho(\zeta_b/\boldsymbol\sigma)\right\}.
\end{aligned}
\label{eq:consistency_reward}
\end{equation}
The function~$\bar\rho$ averages the smooth loss~$\rho(z)=\sqrt{1+z^2}-1$ over the three normalized components. The fixed vector~$\boldsymbol\sigma$ provides per-component normalization scales for the displacement and heading dimensions. The group mean treats all branches symmetrically in the consistency objective. The base tracking rewards encourage command accuracy, while the consistency term reduces variation across loading conditions.

If any branch fails during the window, a shared termination penalty replaces this consistency term for all branches in the group. $\lambda_{\mathrm{con}}$~is scheduled from zero to its final value during training. Window duration, normalization scales, and weight scheduling are specified in Sec.~\ref{sec:impl}.
\section{Experiment}

\begin{table*}[t]
\centering
\caption{Observation and action components under sagittal reflection.}
\label{tab:observation_action}
\small
\begin{tabular}{l c l l}
\hline
\textbf{Component} &
\textbf{Dim.} &
\textbf{Description} &
\textbf{Reflection Transformation} \\
\hline
Base angular velocity $\omega_t^{b}$
& 3 & $(\omega_x,\omega_y,\omega_z)$
& $(-\omega_x,\omega_y,-\omega_z)$ \\
Projected gravity $g_t^{b}$
& 3 & $(g_x,g_y,g_z)$
& $(g_x,-g_y,g_z)$ \\
Velocity command $c_t$
& 3 & $(v_x^{\mathrm{cmd}},v_y^{\mathrm{cmd}},
\omega_z^{\mathrm{cmd}})$
& $(v_x^{\mathrm{cmd}},-v_y^{\mathrm{cmd}},
-\omega_z^{\mathrm{cmd}})$ \\
Joint position $q_t-q^{0}$
& 29 & Joint position relative to the default configuration
& $F_q(q_t-q^{0})$ \\
Joint velocity $\dot q_t$
& 29 & Joint velocity
& $F_q(\dot q_t)$ \\
Previous full-body command $\bar a_{t-1}$
& 29 & Previous command of the 29 joints
& $F_q(\bar a_{t-1})$ \\
Gait phase $\Phi_t$
& 4 & $(\cos\phi_l,\cos\phi_r,\sin\phi_l,\sin\phi_r)$
& $(\cos\phi_r,\cos\phi_l,\sin\phi_r,\sin\phi_l)$ \\
Estimated disturbance wrench $\hat{\mathbf{d}}_t$
& 6 & $(\hat F_x,\hat F_y,\hat F_z,
\hat M_x,\hat M_y,\hat M_z)$
& $(\hat F_x,-\hat F_y,\hat F_z,
-\hat M_x,\hat M_y,-\hat M_z)$ \\
\hline
Base linear velocity $v_t^{b}$
& 3 & $(v_x,v_y,v_z)$
& $(v_x,-v_y,v_z)$ \\
Action $a_t$
& 15 & Leg and waist joint commands
& $F_a(a_t)$ \\
\hline
\end{tabular}
\end{table*}

We evaluate ReACT on the Unitree G1 humanoid robot in both simulation and real-world experiments. The experiments are designed to answer the following questions:
\begin{enumerate}
    \item Can ReACT maintain accurate velocity tracking under varying asymmetric upper-body loads?
    \item How do reflection-consistent modulation and cross-load consistency improve velocity tracking under asymmetric loads?
    \item Can the accurate velocity tracking achieved by ReACT reliably support mapless navigation under asymmetric loads?
\end{enumerate}

\subsection{Implementation Details}
\label{sec:impl}

Policies are trained in Isaac Gym~\cite{makoviychuk2021isaac} with 4096 parallel environments at 50\,Hz for 50K vectorized iterations ($2.048\times10^8$ transitions per training run). FastSAC~\cite{seo2025learning} uses learning rate~$3\times10^{-4}$, discount~$\gamma=0.97$, batch size~8192, and action scale~$\alpha_a=0.25$. The actor encoder is an MLP~(512, 256, 128) with layer normalization and SiLU, followed by two independently parameterized modulation blocks of width~128 and bilinear rank~32. The critic uses two distributional Q-networks~(768, 384, 192), each with 101 categorical atoms over~$[-20,20]$.

The six-dimensional wrench estimator uses a 15-frame proprioceptive history and a group-averaged MLP~(256, 128) whose output respects wrench reflection~$F_d$. It is trained on simulated wrench labels, frozen during policy training, and followed by a first-order low-pass filter~($\tau=0.32$\,s). In an estimator diagnostic over loaded E1 conditions, the asymmetric components ($\hat F_y,\hat M_x,\hat M_z$) have force RMSE ranging from 0.17\,N at 2\,kg to 0.41\,N at 4\,kg and moment RMSE of 0.14--0.56\,Nm. These components achieve 100\% sign agreement with simulation ground truth in these tests.

Velocity tracking rewards use exponential kernels of scale~0.25, with posture, gait-timing, and action-smoothness regularization. Domain randomization covers friction~$[0.5,1.25]$, added base mass~$[-1,3]$\,kg, center-of-mass offsets~$\pm0.05$\,m, actuator gain scaling~$[0.9,1.1]$, and 0.5--6\,kg hand payloads in four loading modes. Evaluation payloads (0.5--4\,kg) are within this range.

For consistency training, 1023 environments form 341 matched triplets (unloaded, left 4\,kg, right 4\,kg), sharing initial states, commands, gait phase, physical parameters, and observation noise on flat terrain; the remaining 3073 use ordinary payload randomization. We set~$T=0.4$\,s, $\boldsymbol\sigma=(0.05,0.05,0.05)$ in (m,\,m,\,rad), and~$c_{\max}=4$. The weight~$\lambda_{\mathrm{con}}$ ramps linearly from~0 to~0.02 between~$10^6$ and~$3\times10^6$ cumulative environment transitions. If any branch terminates within a window, a shared termination penalty replaces the consistency comparison for the whole triplet. All methods receive the same prescribed upper-body reference during evaluation. FALCON uses its released checkpoint~\cite{zhang2025falcon} and native observations. We omit a quantitative comparison with concurrent ADAPT~\cite{lyu2026adaptanalyticaldisturbanceawarepolicy}, whose code and checkpoints were not publicly available during our experiments (see Sec.~\ref{sec:rw_disturbance} for a methodological comparison).

\subsection{Velocity Tracking Performance Evaluation}
\label{sec:velocity_tracking}

Table~\ref{tab:tracking_protocols} defines three protocols: straight walking~(E1), constant turning~(E2), and dynamic commands~(E3), all with~$v_y^{\mathrm{cmd}}=0$. For each trial, we compute tracking MAE
\begin{equation}
E_i=\operatorname{mean}_t\!\left[|\varepsilon_i(t)|\right],
\quad i\in\{v_x,v_y,\omega\},
\label{eq:eval_tracking_mae}
\end{equation}
where~$\varepsilon_i$ is the signed tracking residual (measured minus commanded), and~$\operatorname{mean}_t$ averages over the evaluation window. For E1, we additionally quantify persistent directional error using the signed bias
\begin{equation}
B_i=\operatorname{mean}_t\!\left[\varepsilon_i(t)\right],
\quad i\in\{v_y,\omega\}.
\label{eq:signed_bias}
\end{equation}
We also evaluate actuator effort in all three protocols using the normalized torque cost
\begin{equation}
J_\tau=\operatorname{mean}_t\!\left[
\frac{1}{15}\sum_{j=1}^{15}
\left(\frac{\tau_j(t)}{\tau_j^{\mathrm{lim}}}\right)^2
\right],
\label{eq:eval_torque_cost}
\end{equation}
using applied torques at physics substeps and torque limits for the 12 leg and 3 waist joints.
E1/E2 use 10\,s rollouts scored over~$[2,10)$\,s; E3 scores the full~$[0,37)$\,s fixed command sequence.
All rollouts follow an excluded 1.3\,s zero-command bootstrap. Each setting uses three trials with different initial timings of the prescribed upper-body motion, matched across methods, yielding 102, 120, and 30 rollouts per checkpoint for E1, E2, and E3, respectively. Per-trial metrics are averaged within each checkpoint over the conditions in each table row. For ReACT and its variants, we report the mean and sample standard deviation~(SD) across two independent training seeds.

\begin{table*}[t]
\centering
\caption{Velocity tracking under asymmetric hand loads.
ReACT: mean $\pm$ sample SD over two training seeds.
FALCON: one released checkpoint. Lower is better.
Best means are bold.}
\label{tab:velocity_tracking}
\label{tab:tracking_protocols}
\begingroup
\small
\setlength{\tabcolsep}{2.5pt}
\renewcommand{\arraystretch}{1.05}
\begin{tabular}{@{}lrrrrrrrr@{}}
\hline\hline
& \multicolumn{4}{c}{FALCON}
& \multicolumn{4}{c}{ReACT (Ours)} \\
\cline{2-5}\cline{6-9}
Protocol
& $E_{v_x}$ & $E_{v_y}$ & $E_{\omega}$ & $J_\tau$
& $E_{v_x}$ & $E_{v_y}$ & $E_{\omega}$ & $J_\tau$ \\
& (m/s) & (m/s) & (rad/s) &
& (m/s) & (m/s) & (rad/s) & \\
\hline
E1: straight
& 0.1033 & 0.1212 & 0.5115 & 0.03744
& $\mathbf{0.0632} \pm 0.0055$
& $\mathbf{0.0824} \pm 0.0032$
& $\mathbf{0.0924} \pm 0.0181$
& $\mathbf{0.02236} \pm 0.00131$ \\
E2: turning
& 0.1086 & 0.1401 & 0.5687 & 0.03913
& $\mathbf{0.0579} \pm 0.0070$
& $\mathbf{0.0932} \pm 0.0100$
& $\mathbf{0.1049} \pm 0.0237$
& $\mathbf{0.02389} \pm 0.00191$ \\
E3: Dynamic A
& 0.1204 & 0.1289 & 0.5574 & 0.03311
& $\mathbf{0.0567} \pm 0.0015$
& $\mathbf{0.0872} \pm 0.0072$
& $\mathbf{0.0877} \pm 0.0218$
& $\mathbf{0.01957} \pm 0.00160$ \\
E3: Dynamic B
& 0.1184 & 0.1277 & 0.5580 & 0.03292
& $\mathbf{0.0562} \pm 0.0014$
& $\mathbf{0.0867} \pm 0.0067$
& $\mathbf{0.0872} \pm 0.0212$
& $\mathbf{0.01952} \pm 0.00154$ \\
\hline\hline
\end{tabular}
\par\smallskip
\parbox{\textwidth}{\footnotesize
\textbf{E1:} $v_x^{\mathrm{cmd}}\in\{0.8,1.0\}$\,m/s,
$\omega_z^{\mathrm{cmd}}=0$; unloaded or 0.5--4\,kg
on either hand (0.5\,kg steps). Both speeds are pooled.\par
\textbf{E2:} Same forward commands,
$\omega_z^{\mathrm{cmd}}\in\{\pm0.3,\pm0.6\}$\,rad/s;
unloaded or 2 or 4\,kg on either hand.\par
\textbf{E3:} Two fixed sequences A/B with nominal
$v_x^{\mathrm{cmd}}\in\{0,0.8,1.0\}$\,m/s and
$\omega_z^{\mathrm{cmd}}\in\{0,\pm0.3,\pm0.6\}$\,rad/s,
differing only in yaw-command signs; loads as in E2.
Trial conditions are matched across methods.}
\endgroup
\end{table*}

ReACT has lower mean tracking MAE and normalized torque cost than FALCON in E1, E2, and both E3 sequences (Table~\ref{tab:velocity_tracking}). Forward, lateral, and yaw-rate MAE decrease by 38.8--52.9\%, 32.0--33.5\%, and 81.6--84.4\%, respectively, while torque cost decreases by 38.9--40.9\%. The tracking advantage holds for both evaluated ReACT seeds, and sequences A/B show similar performance.

\begin{figure}[t]
    \centering
    \includegraphics[width=0.80\columnwidth]{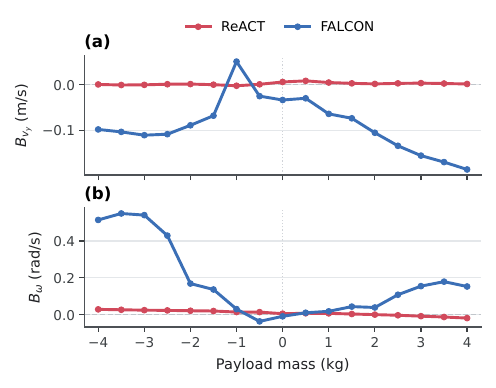}
    \caption{Signed directional bias versus unilateral
    payload mass at $v_x^{\mathrm{cmd}}=1.0$\,m/s,
    using an illustrative checkpoint of each method.
    Negative mass denotes left-hand loading, positive
    denotes right-hand loading.}
    \label{fig:bias_mass}
\end{figure}

\begin{figure*}[t]
    \centering
    \includegraphics[width=0.95\textwidth]{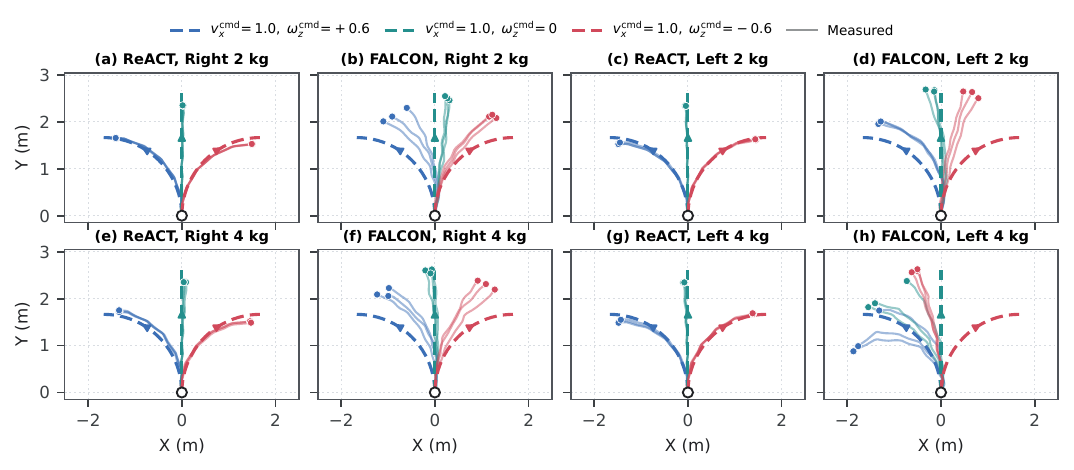}
    \caption{Planar trajectories for straight walking and
    quarter turns at 1.0\,m/s. Top: 2\,kg; bottom: 4\,kg.
    Dashed lines show commanded paths; solid lines show
    measured trajectories from three evaluation trials
    for an illustrative checkpoint of each method.}
    \label{fig:trajectory}
\end{figure*}

Figs.~\ref{fig:bias_mass} and~\ref{fig:trajectory} provide checkpoint-specific illustrations of these behaviors. Both methods complete the corresponding evaluation rollouts without falls. In this checkpoint-specific E1 evaluation at 0.8\,m/s, FALCON's unloaded yaw-rate MAE is 0.199\,rad/s and its pooled loaded MAE is 0.518\,rad/s. Across loaded E1 conditions, its signed biases reach $|B_{v_y}|=0.186$\,m/s and~$|B_\omega|=0.598$\,rad/s. For the ReACT checkpoint used in these illustrations, $|B_{v_y}|<0.009$\,m/s and $|B_\omega|<0.029$\,rad/s across E1 conditions, reducing the maximum bias by over an order of magnitude. Two-seed bias-range statistics are reported in Sec.~\ref{sec:ablation}.

Fig.~\ref{fig:trajectory} shows straight walking and quarter turns at 1.0\,m/s. ReACT closely follows the commanded paths at both payload magnitudes and loading sides, whereas FALCON exhibits side-dependent deviations. The navigation policy in Sec.~\ref{sec:high_level_task} commands forward velocity and yaw rate without a lateral channel. Persistent lateral bias therefore requires corrective heading changes and can reduce obstacle clearance.

\begin{table*}[t]
\centering
\caption{Component ablations and payload-randomization
control. Best means are bold.}
\label{tab:ablation}
\begingroup
\footnotesize
\setlength{\tabcolsep}{2pt}
\renewcommand{\arraystretch}{1.04}
\begin{tabular*}{\textwidth}{@{\extracolsep{\fill}}l*{12}{c}@{}}
\hline\hline
& \multicolumn{4}{c}{E1: straight walking}
& \multicolumn{4}{c}{E2: constant turning}
& \multicolumn{4}{c}{E3: dynamic commands} \\
\cline{2-5}\cline{6-9}\cline{10-13}
Method
& $E_{v_x}$ & $E_{v_y}$ & $E_{\omega}$ & $J_\tau$
& $E_{v_x}$ & $E_{v_y}$ & $E_{\omega}$ & $J_\tau$
& $E_{v_x}$ & $E_{v_y}$ & $E_{\omega}$ & $J_\tau$ \\
\hline
w/o Consistency
& 0.2371 & 0.1747 & 0.2007 & 0.02042 & 0.2823 & 0.1679 & 0.2622 & 0.02117 & 0.2673 & 0.1087 & 0.1622 & 0.02251 \\[-1.5pt]
& {\scriptsize $\pm 0.0068$} & {\scriptsize $\pm 0.0193$} & {\scriptsize $\pm 0.0299$} & {\scriptsize $\pm 0.00003$} & {\scriptsize $\pm 0.0096$} & {\scriptsize $\pm 0.0048$} & {\scriptsize $\pm 0.0362$} & {\scriptsize $\pm 0.00021$} & {\scriptsize $\pm 0.0180$} & {\scriptsize $\pm 0.0068$} & {\scriptsize $\pm 0.0158$} & {\scriptsize $\pm 0.00248$} \\[1pt]
w/o Modulation
& 0.2338 & 0.1937 & 0.2523 & 0.02087 & 0.3102 & 0.2054 & 0.3135 & 0.02227 & 0.2562 & 0.1611 & 0.2068 & 0.02146 \\[-1.5pt]
& {\scriptsize $\pm 0.0495$} & {\scriptsize $\pm 0.0040$} & {\scriptsize $\pm 0.0136$} & {\scriptsize $\pm 0.00132$} & {\scriptsize $\pm 0.0716$} & {\scriptsize $\pm 0.0119$} & {\scriptsize $\pm 0.0053$} & {\scriptsize $\pm 0.00062$} & {\scriptsize $\pm 0.0699$} & {\scriptsize $\pm 0.0284$} & {\scriptsize $\pm 0.0349$} & {\scriptsize $\pm 0.00163$} \\[1pt]
ReACT-DR
& 0.3383 & 0.1897 & 0.2204 & 0.02023 & 0.3892 & 0.1757 & 0.2955 & \textbf{0.01973} & 0.3677 & 0.1384 & 0.1916 & \textbf{0.01808} \\[-1.5pt]
& {\scriptsize $\pm 0.0057$} & {\scriptsize $\pm 0.0140$} & {\scriptsize $\pm 0.0530$} & {\scriptsize $\pm 0.00016$} & {\scriptsize $\pm 0.0197$} & {\scriptsize $\pm 0.0072$} & {\scriptsize $\pm 0.0177$} & {\scriptsize $\pm 0.00007$} & {\scriptsize $\pm 0.0270$} & {\scriptsize $\pm 0.0004$} & {\scriptsize $\pm 0.0239$} & {\scriptsize $\pm 0.00025$} \\[1pt]
Equiv.\ Conditioning
& 0.2407 & 0.1508 & 0.2127 & \textbf{0.01960} & 0.2861 & 0.1610 & 0.2956 & 0.02100 & 0.2386 & 0.1337 & 0.1835 & 0.02067 \\[-1.5pt]
& {\scriptsize $\pm 0.0423$} & {\scriptsize $\pm 0.0152$} & {\scriptsize $\pm 0.0208$} & {\scriptsize $\pm 0.00124$} & {\scriptsize $\pm 0.0378$} & {\scriptsize $\pm 0.0263$} & {\scriptsize $\pm 0.0069$} & {\scriptsize $\pm 0.00172$} & {\scriptsize $\pm 0.0139$} & {\scriptsize $\pm 0.0224$} & {\scriptsize $\pm 0.0014$} & {\scriptsize $\pm 0.00075$} \\[1pt]
ReACT
& \textbf{0.0632} & \textbf{0.0824} & \textbf{0.0924} & 0.02236 & \textbf{0.0579} & \textbf{0.0932} & \textbf{0.1049} & 0.02389 & \textbf{0.0565} & \textbf{0.0870} & \textbf{0.0874} & 0.01954 \\[-1.5pt]
& {\scriptsize $\pm 0.0055$} & {\scriptsize $\pm 0.0032$} & {\scriptsize $\pm 0.0181$} & {\scriptsize $\pm 0.00131$} & {\scriptsize $\pm 0.0070$} & {\scriptsize $\pm 0.0100$} & {\scriptsize $\pm 0.0237$} & {\scriptsize $\pm 0.00191$} & {\scriptsize $\pm 0.0014$} & {\scriptsize $\pm 0.0070$} & {\scriptsize $\pm 0.0215$} & {\scriptsize $\pm 0.00157$} \\[1pt]
\hline\hline
\end{tabular*}
\endgroup
\end{table*}

\begin{figure*}[t]
    \centering
    \includegraphics[width=\textwidth]{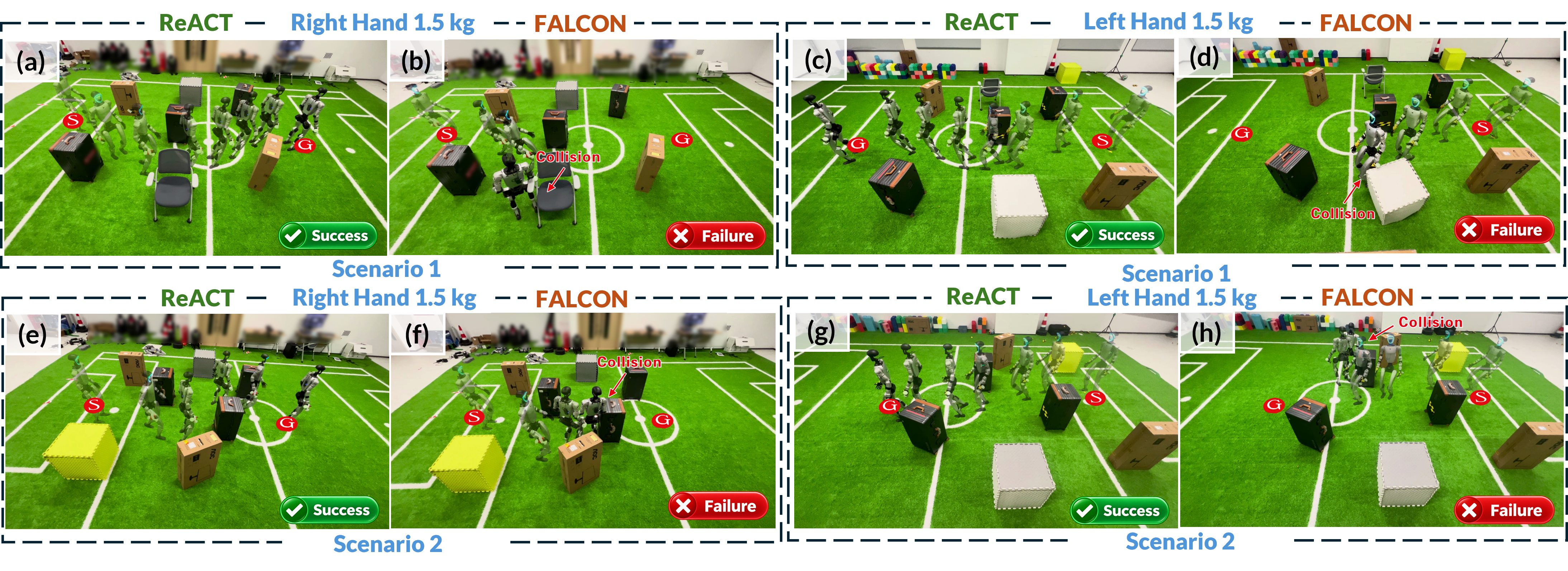}
    \caption{Real-world navigation comparison between
    ReACT and FALCON under unilateral hand loads of
    1.5\,kg. S and G denote the start and goal,
    respectively.}
    \label{fig:real-world}
\end{figure*}

\begin{figure}[t]
    \centering
    \begin{minipage}[t]{0.35\linewidth}
        \centering
        \includegraphics[width=\linewidth]{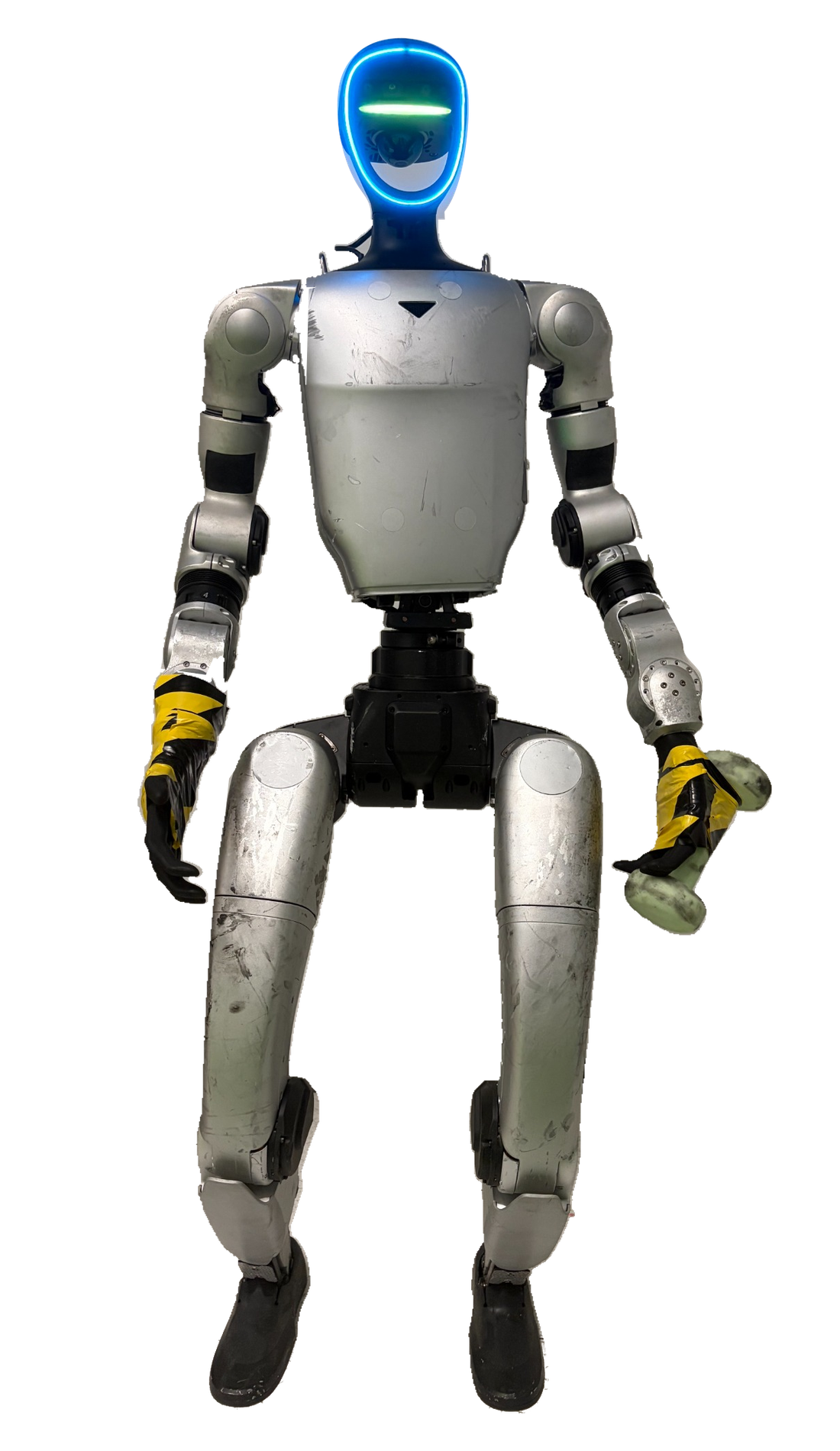}
        \par\smallskip
        {\small (a)}
    \end{minipage}%
    \hspace{0.02\linewidth}%
    \begin{minipage}[t]{0.35\linewidth}
        \centering
        \includegraphics[width=\linewidth]{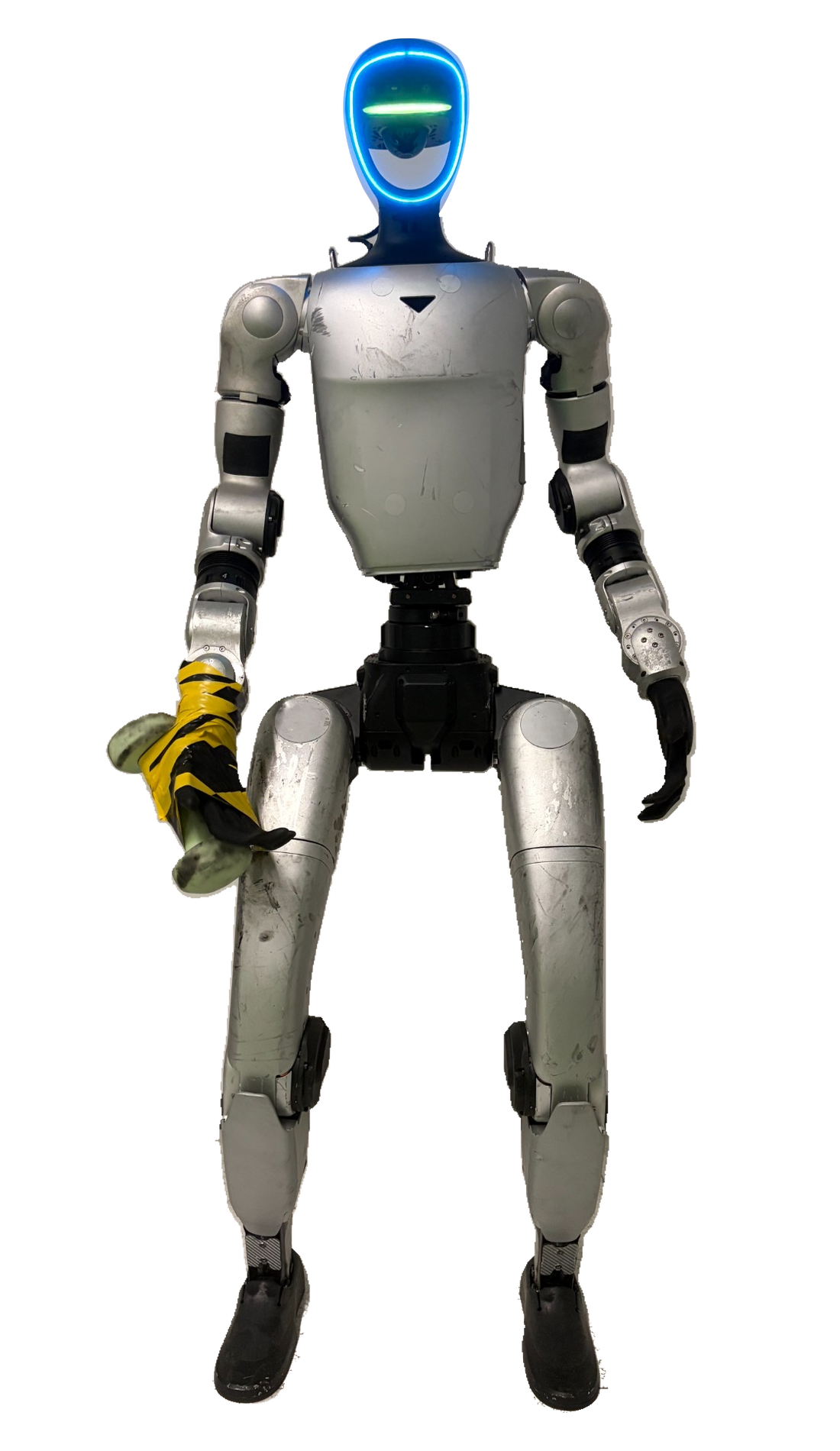}
        \par\smallskip
        {\small (b)}
    \end{minipage}
    \caption{Upper-body load configurations of the
    Unitree G1 used in real-world experiments:
    (a)~a 1.5\,kg payload in the left hand with the
    right hand unloaded;
    (b)~a 1.5\,kg payload in the right hand with
    the left hand unloaded.}
    \label{fig:robot_info}
\end{figure}

\subsection{Ablation Studies}
\label{sec:ablation}

To evaluate the contributions of the proposed components, we compare ReACT with four variants:
\begin{enumerate}
    \item \textbf{ReACT w/o Consistency:}
    We disable cross-load tracking consistency while retaining reflection-consistent disturbance modulation. The payload exposure remains unchanged.

    \item \textbf{ReACT w/o Modulation:}
    We replace reflection-consistent disturbance modulation with standard MLP wrench conditioning, while retaining cross-load tracking consistency and the same wrench information.

    \item \textbf{ReACT-DR:}
    We disable cross-load tracking consistency and independently randomize hand payloads while retaining reflection-consistent disturbance modulation. The per-environment payload marginal matches that of the paired configuration, but cross-branch correlation is removed.

    \item \textbf{Equiv. Conditioning:}
    We retain the group-averaged equivariant encoder and cross-load tracking consistency, but replace the modulation blocks with direct wrench concatenation at the encoder input. This variant isolates the contribution of modulation beyond the equivariant prior.
\end{enumerate}

All five methods are trained with the same two random seeds and evaluated at the 50K checkpoint. ReACT, w/o Consistency, w/o Modulation, and Equiv.\ Conditioning share the same paired and ordinary payload exposure.
Table~\ref{tab:ablation} reports means with sample SDs below over two independent training seeds, pooling both E1 speeds and E3 sequences A/B. Units follow Table~\ref{tab:velocity_tracking}.

ReACT achieves the lowest mean forward, lateral, and yaw-rate errors across E1--E3, with the tracking advantage holding for both evaluated seeds. Equiv.\ Conditioning improves mean lateral tracking over w/o Modulation in all three protocols and mean forward tracking in E2/E3, but remains worse than ReACT. Relative to Equiv.\ Conditioning, ReACT reduces mean forward velocity MAE by 73.7--79.8\% and yaw-rate MAE by 52.4--64.5\%, supporting the benefit of wrench modulation beyond direct equivariant conditioning.

With modulation retained, ReACT reduces mean yaw-rate MAE by 46.1--60.0\% relative to w/o Consistency across E1--E3, supporting consistency training within the ReACT architecture. Relative to w/o Consistency, ReACT-DR has higher mean forward and lateral errors in all three protocols, despite the lowest mean torque cost in E2/E3. Equiv.\ Conditioning has the lowest mean torque cost in E1. Lower torque cost therefore does not consistently indicate better tracking in these comparisons.

To assess action correspondence under observation reflection, we evaluate the deterministic policy on a shared set~$\mathcal D$ of 9,250 held-out actor observations:
\begin{equation}
R_q=\alpha_a\operatorname{mean}_{o\in\mathcal D}
\!\left[\left\|\pi_\theta(F_o(o))-F_a\pi_\theta(o)
\right\|_2\right],
\label{eq:reflection_residual}
\end{equation}
where~$\alpha_a=0.25$\,rad per action unit converts the policy outputs to residual joint-position targets, and the $L_2$ norm covers the 15 actor-controlled joints. All variants with equivariant architectures (ReACT, w/o Consistency, ReACT-DR, and Equiv.\ Conditioning) yield numerically zero residuals in both seeds. Standard MLP wrench conditioning (w/o Modulation) yields $R_q=0.0695\pm0.0103$\,rad (mean $\pm$ sample SD), indicating nonzero reflection error.

For each checkpoint, we average signed biases over the three trials per condition and compute their range $\Delta B_i$ over unloaded, left 4\,kg, and right 4\,kg E1 conditions. Values below are mean $\pm$ sample SD over two seeds, listed at 0.8 and 1.0\,m/s, respectively. ReACT achieves $\Delta B_{v_y}=0.0114\pm0.0048$ and $0.0058\pm0.0005$\,m/s, versus $0.2545\pm0.0938$ and $0.3049\pm0.1303$\,m/s for w/o Consistency. The four variants' mean ranges are 0.1075--0.2545\,m/s at 0.8\,m/s and 0.1263--0.3049\,m/s at 1.0\,m/s. Equiv.\ Conditioning retains larger lateral bias ranges ($0.1899\pm0.0933$ and $0.2488\pm0.1299$\,m/s) despite sharing the reflection constraint and consistency objective. This supports modulation beyond direct equivariant conditioning under the shared consistency objective.

ReACT's yaw-bias ranges are $0.0265\pm0.0146$ and $0.0338\pm0.0192$\,rad/s, versus $0.1610\pm0.0412$ and $0.1976\pm0.0531$\,rad/s for w/o Consistency. Both bias ranges are lower with consistency training within ReACT. At 1.0\,m/s, w/o Modulation has a lower mean yaw-bias range ($0.0284\pm0.0273$\,rad/s). Separately, over the full E1 protocol (both speeds and all payload conditions), its yaw-rate MAE is 2.7 times that of ReACT (Table~\ref{tab:ablation}).

\subsection{Real-World Evaluation}
\label{sec:high_level_task}

To quantify velocity tracking on the physical robot, we record odometry and IMU measurements during 1\,m/s turning under 1.5\,kg hand loads. Table~\ref{tab:real_tracking} reports the MAE over stable 1.5\,s windows. ReACT has lower forward and yaw-rate MAE, but higher lateral MAE in these turning windows.

\begin{table}[t]
\centering
\caption{Real-robot velocity tracking MAE during
1\,m/s turning with~$|\omega_z^{\mathrm{cmd}}|
\ge0.8$\,rad/s under 1.5\,kg hand loads.}
\label{tab:real_tracking}
\small
\setlength{\tabcolsep}{6pt}
\renewcommand{\arraystretch}{1.15}
\begin{tabular}{lccc}
\hline\hline
Method
& $E_{v_x}$ (m/s)
& $E_{v_y}$ (m/s)
& $E_\omega$ (rad/s) \\
\hline
ReACT & \textbf{0.098} & 0.159 & \textbf{0.525} \\
FALCON & 0.202 & \textbf{0.135} & 0.696 \\
\hline\hline
\end{tabular}
\end{table}

We evaluate real-world mapless navigation under asymmetric hand loads using the same DRL-DCLP navigation policy~\cite{zhang2025drl} for both locomotion controllers. It commands forward velocity and yaw rate.

As shown in Fig.~\ref{fig:real-world}, the humanoid robot was evaluated in two real-world scenarios under 1.5\,kg unilateral hand loads~(Fig.~\ref{fig:robot_info}). Each configuration combines a scenario and a loading side~(R:~right hand, L:~left hand, 1/2:~scenario index), with 10 independent trials per configuration and 40 trials per method. In each trial, the robot navigates from a start position~$S$ to a goal~$G$ while avoiding obstacles. A trial is successful when the robot reaches within 0.35\,m of the goal without collision within 60\,s. Collision or timeout is counted as failure.

\begin{table}[t]
\centering
\caption{Real-world navigation under 1.5\,kg
asymmetric hand loads. Path length and motion time are
mean $\pm$ standard deviation over successful trials.}
\label{tab:nav_results}
\begingroup
\small
\setlength{\tabcolsep}{3.5pt}
\renewcommand{\arraystretch}{1.15}
\begin{tabular}{lcccc}
\hline\hline
& \multicolumn{2}{c}{Success}
& \multicolumn{2}{c}{ReACT metrics} \\
\cline{2-3}\cline{4-5}
Config.
& ReACT & FALCON
& Path~(m) & Time~(s) \\
\hline
R1 & \textbf{10/10} & 0/10
   & 4.17$\pm$0.23 & 11.39$\pm$1.69 \\
L1 & \textbf{10/10} & 0/10
   & 4.36$\pm$0.53 & 14.51$\pm$4.32 \\
R2 & \textbf{10/10} & 0/10
   & 4.63$\pm$0.43 & 27.67$\pm$3.80 \\
L2 & \textbf{10/10} & 0/10
   & 2.91$\pm$0.30 & 10.29$\pm$2.72 \\
\hline
Total & \textbf{40/40} & 0/40
      & -- & -- \\
\hline\hline
\end{tabular}
\endgroup
\end{table}

Table~\ref{tab:nav_results} reports the navigation results. ReACT succeeds in all 40 trials, while all 40 FALCON trials fail due to collisions. These results demonstrate collision-free goal reaching with ReACT under the fixed navigation policy in the tested scenarios.
\section{Conclusion}

In this paper, we presented ReACT, a framework for accurate velocity tracking under persistent asymmetric upper-body loads on humanoid robots. We focused on the transport phase of humanoid loco-manipulation, where unilateral loads can introduce persistent lateral and yaw-rate biases even during stable walking. The resulting position and heading deviations can compromise obstacle avoidance and goal reaching. ReACT combines reflection-consistent disturbance modulation with cross-load tracking consistency training to adapt compensation to the estimated wrench and reduce tracking variation across loads. Simulation experiments showed improved velocity tracking and reduced directional bias. In real-world experiments on a Unitree G1 humanoid robot, ReACT achieved collision-free goal reaching in all 40 navigation trials under unilateral hand loads with a fixed navigation policy. These results demonstrate ReACT's practical value for humanoid robot navigation during object transport. Future work will extend ReACT to dynamic load changes during manipulation tasks and evaluate tracking across a broader range of real-world commands and payloads.

\bibliographystyle{IEEEtran}
\bibliography{Bibliography/mybib}

\end{document}